\documentclass[final]{cvpr}
\usepackage{times}
\usepackage{epsfig}
\usepackage{graphicx}
\usepackage{amsmath}
\usepackage{amssymb}
\usepackage{multirow}
\usepackage{url}

\usepackage[pagebackref=true,breaklinks=true,colorlinks,bookmarks=false]{hyperref}

\def\cvprPaperID{7701} 
\def\confYear{CVPR 2021}
\begin{document}
\title{Pixel Codec Avatars}
\author{Shugao Ma \qquad Tomas Simon \qquad Jason Saragih \qquad Dawei Wang \\ 
Yuecheng Li \qquad Fernando De La Torre \qquad Yaser Sheikh\\
Facebook Reality Labs Research\\
{\tt\small \{shugao, tsimon, jsaragih, dawei.wang, yuecheng.li, ftorre, yasers\}@fb.com}
}

\maketitle

\begin{abstract}
 Telecommunication with photorealistic avatars in virtual or augmented reality is a promising path for achieving authentic face-to-face communication in 3D over remote physical distances. In this work, we present the Pixel Codec Avatars (PiCA): a deep generative model of 3D human faces that achieves state of the art reconstruction performance while being computationally efficient and adaptive to the rendering conditions during execution. Our model combines two core ideas: (1) a fully convolutional architecture for decoding spatially varying features, and (2) a rendering-adaptive per-pixel decoder. Both techniques are integrated via a dense surface representation that is learned in a weakly-supervised manner from low-topology mesh tracking over training images. We demonstrate that PiCA improves reconstruction over existing techniques across testing expressions and views on persons of different gender and skin tone. Importantly, we show that the PiCA model is much smaller than the state-of-art baseline model, and makes multi-person telecommunicaiton possible: on a single Oculus Quest 2 mobile VR headset, 5 avatars are rendered in realtime in the same scene.    
\end{abstract}

\section{Introduction}
\label{sec:intro}
\begin{figure}[t!]
    \centering
    \includegraphics[height=0.39\textwidth]{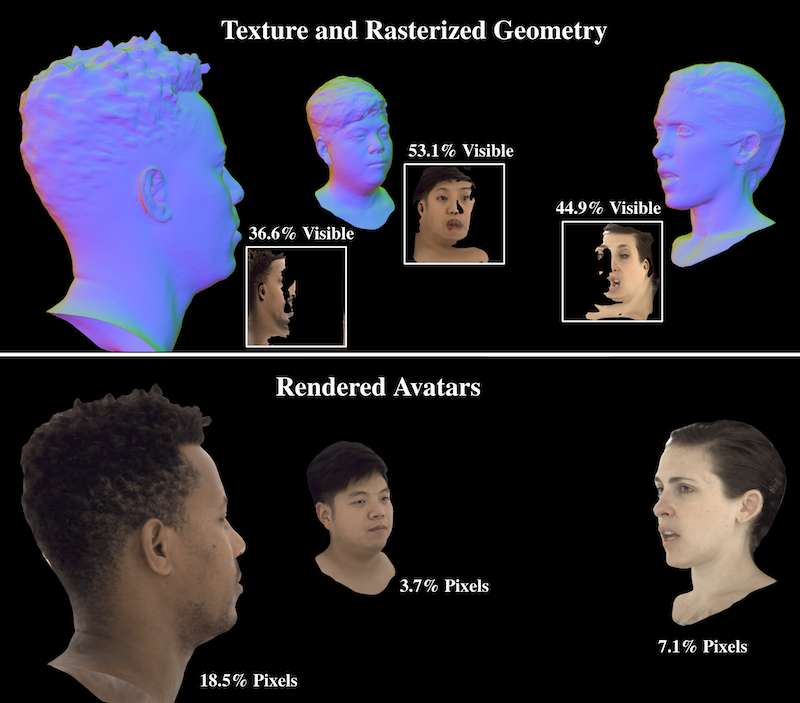}
    
    \caption{An multi-person configuration for teleconference in VR. At normal interpersonal distances~\cite{social_distance}, the head occupies only a subset of pixels in the display, where the amount of coverage largely depends on distance to the viewer. Roughly half of the head is not visible from any viewing angle due to self occlusion. Our method avoids wasting computation on areas that do not directly contribute to the final image. In first row we show the generated and rasterized geometry, along with texture maps showing visible pixels from the corresponding views; in the second row we show the rendered avatars and the percentage of pixels they cover over the entire image.}
    \label{fig:intro}
\end{figure}

Photorealistic Telepresence in Virtual Reality (VR) as proposed in \cite{Stephen18, Shih-En19}, describes a technology for enabling authentic communication over remote distances that each communicating party {\it feels the genuine co-location presence} of the others. At the core of this technology is the {\em Codec Avatar}, which is a high fidelity animatable human face model, implemented as the decoder network of a Variational AutoEncoder (VAE). Imagine a two-way communication setting. At the transmitter end, an encoding process is performed: cameras mounted on transmitter's VR headset capture partial facial images and an encoder model encodes the captured images into latent code of the decoder in realtime. At the receiver end a decoding process is performed: upon receiving the latent code over the internet, the decoder decodes the avatar's geometry and appearance so that the transmitter's realtime photorealistic face can be rendered onto the VR display.

Multi-person communication via Photorealistic VR Telepresence will enable applications that are in great need in the modern society, such as family re-union over far physical distances in which each member genuinely feels the co-location presences of the others, or collaboration in remote working where team members can effectively communicate face-to-face in 3D. However, rendering with the decoder model proposed in \cite{Stephen18} does not scale well with the number of communicating parties. Specifically, a full texture of fixed resolution 1K$\times$1K is decoded at each frame despite the distance of the avatar to the viewer and visibility of different facial regions. This leads to significant waste of computation when the avatar is far away, for which case the rendered avatar only consists a small number of pixels (Fig. \ref{fig:intro}), resulting in a large number of pixels in the decoded texture map unused. Also, most of the time half of the head is not visible due to self-occlusion, so the pixels in the decoded texture map for the occluded part are also unused. For a 2K display such as the one in Quest2, rendering more than 4 avatars amounts to computing more pixels than that of the display. This is obviously limiting, e.g. family re-union of more than 4 persons or team collaboration of more than 4 members are common place.  

To solve this issue and scale the rendering to the number of persons in the VR telepresence, we should compute only the visible pixels, thus upper bounding the computation by the number of pixels of the display. Recent works in neural rendering such as the {\em defferred neural rendering}\cite{Justus19}, the {\em neural point-based graphics}\cite{Aliev19}, the {\em implicit differentiable rendering} \cite{Lior20}, use neural network to compute pixel values in the screen space instead of the texture space thus computing only visible pixels. However, in all these works, either a static scene is assumed, or the viewing distance and perspective are not expected to be entirely free in the 3D space. However, for telepresence, the ability to animate the face in realtime and render it from any possible viewing angle and distance is crucial.

In this paper, we present Pixel Codec Avatars (PiCA) that aims to achieve efficient and yet high fidelity dynamic human face rendering that is suitable for multi-person telepresence in VR on devices with limited compute. To avoid wasteful computation in areas of the face that do not contribute to the final rendering, PiCA employs per-pixel decoding only in areas of the image covered by a rasterization of the geometry. Similar to recent advances in implicit neural rendering~\cite{mildenhall2020nerf,sitzmann2019siren,tancik2020fourfeat}, this decoder relies on a rich face-centric position encoding to produce highly detailed images. We employ two strategies to generate such encodings efficiently. First, we make use of the spatially-shared computation of convolutional networks in texture space to produce spatially varying expression- and view-specific codes at a reduced resolution (256$\times$256). This is complemented by a pre-computed high resolution (1K$\times$1K) \emph{learned non-parametric positional encoding}, that is jointly rasterized into screen space similarly to~\cite{Justus19}. To achieve an even higher resolution result, we further compliment the signal with 1D positional encodings at 10K resolution, independently for the horizontal and vertical dimensions of the texture domain. Together, these maps enable the modeling of sharp spatial details present in high resolution facial images. Because the best encoding values for the UV coordinates are directly learned from data, a low 8-dimensional encoding is sufficient to recover high frequencies. This is in contrast to existing positional encoding schemes (\eg \cite{mildenhall2020nerf}) that achieve high details using sinusoidal functions, but require increasing the dimensionality by 20$\times$, with corresponding computational costs. Secondly, in contrast to other works such as \cite{Justus19, Aliev19, Lior20}, we do not employ convolutions in screen space, but instead apply a shallow MLP at each contributing pixel. This has the advantage of avoiding visual artifacts during motion and stereo inconsistencies, as well as challenges in generalizing to changes in scale, rotation and perspective, all of which are common in interactive immersive 3D media. 

Our other main insight is that the complexity of view-dependent appearance in prior work stems mostly from inadequate geometric models of the face. Recent work into implicit scene modeling (i.e. NeRF~\cite{mildenhall2020nerf}) has demonstrated that complex view dependent effects such as specularity can be adequately modeled using a shallow network given good estimates of the scene's geometry. Inspired by these results, our construction involves a variational geometry decoder that is learned in a self-supervised manner, using image and depth reconstruction as a supervisory signal. The resulting mesh acquired from this decoder contains more accurate geometry information, substantially simplifying the view-dependent texture generation task, allowing for the use of lightweight pixel-wise decoding.  

\vspace{0.5em}
\noindent
\textbf{Contributions}:
Our contributions are as follows:
\begin{itemize}
    \item We propose {\em Pixel Codec Avatar}, a novel light weight representation that decodes only the visible pixels on the avatar's face in the screen space towards enabling high fidelity facial animation on compute-constrained platforms such as mobile VR headsets.
    \item We make the two major technical innovations to achieve high quality decoding with a small model: {\em learned} positional encoding functions and fully convolutional dense mesh decoder trained in a weakly-supervised fashion.
\end{itemize}

\section{Related Works}
\label{sec:related}
\subsection{Deep 3D Morphable Face Models}
3D Morphable Face Models (3DMFM) are a generative model for 3D human faces. The early works explore ways to represent human facial deformations and appearance with linear subspace representations.
Blanz \etal \cite{Blanz99} models 
shape and texture of human faces as vector spaces and generates new faces and expressions as linear combinations of 
the prototype vectors. Since then, blendshape models have been extensively studied and applied in animation - 
\cite{Lewis2014PracticeAT} provides a good overview of such methods. 
To achieve highly expressive models, a large number of blendshapes need to be manually created and refined, 
\eg the character of Gollum in the movie {\em Lord of the Rings} had 946 blendshapes taking over a year's time to create \cite{Raitt04}. 

In recent years, deep learning techniques, especially generative models such as Variational Auto-Encoder (VAE)~\cite{VAE} and 
Generative Adversarial Networks (GAN)~\cite{GAN} have been actively studied for creating non-linear 3D Morphable Face Model analogues. Tewari \etal~\cite{tewari17MoFA} propose a deep convolutional architecture for monocular face reconstruction, learned from morphable models. 
Lombardi \etal~\cite{Stephen18} propose to jointly model face shape and appearance with a VAE: the encoder encodes the facial mesh and texture into latent code with 
fully connected layers and convolutional layers respectively, and the decoder decodes back the facial mesh and view direction conditioned texture with
fully connected layers and transposed convolutional layers respectively. This model has been referred to as a {\em Codec Avatar} by several subsequent 
works \cite{Shih-En19,Chu2020,Gabriel20, richard2020audio} which animate this model using visual and/or audio sensory data.  
Tran \etal \cite{Tran_2018_CVPR} also use an autoencoder to model geometry and texture, but train the model from unconstrained face images using a rendering loss. 
Bagautdinov \etal \cite{Bagautdinov_2018_CVPR} uses a compositional VAE to model details of different granularities of facial geometry via multiple layers of hidden variables. 
Ranjan \etal \cite{Ranjan_2018_ECCV} directly applies mesh convolution to build a mesh autoencoder 
while Zhou \etal \cite{Zhou_2019_CVPR} extends this idea and jointly models texture and geometry with mesh convolution,
leading to a colored mesh decoder. 

Generative Adversarial Network (GAN) is also explored. Among the first works that use GAN models to build 3DMFM, Slossberg \etal \cite{Slossberg_2018_ECCV_Workshops} build a GAN model that generates realistic 2D texture image as well as coefficients of a PCA based facial mesh model. 
Abrevaya \etal \cite{Abrevaya_2019_ICCV} maps mesh to geometry image (\ie equivalent to position map in this paper) and builds a GAN model of the mesh
that has decoupled expression and identity codes, and the decoupling is achieved with auxilary expression and identity classification tasks during training. 
Shamai \etal \cite{Shamai2019} also maps mesh into geometry image and builds GAN models using convolutional layers for both geometry and texture. 
Cheng \etal \cite{cheng2019meshgan} proposes GAN model of facial geometry with mesh convolution.

The most distinctive feature of PiCA against the previous 3DMFM is that the pixel decoder decodes color at each pixel given underlying geometry that is generated and rasterized to screen space, hence
adaptive resolution and computational cost is achieved. In contrast, in all previous methods, texture is either modeled as a 2D texture map \cite{Stephen18, Tran_2018_CVPR, Slossberg_2018_ECCV_Workshops} thus fixing the 
output resolution, or is modeled at mesh vertices \cite{Zhou_2019_CVPR, Shamai2019}, thus mesh density determines the rendering resolution. Another advantage is that our method explicitly models the correlation between geometry and texture in the per-object decoding step, which is lacking in most previous 3D DFMM models. 

\subsection{Neural Rendering}
Our method is also related to recent works on Neural Rendering and \cite{tewari2020state} provides a good survey of recent progress in this direction. 
In particular, Thies \etal \cite{Justus19} propose deferred neural rendering with a neural texture, which in spirit is close to our work: 
neural textures, \ie a feature output from a deep neural net, is rasterized to screen space and another neural net, \ie the neural renderer, 
computes colors from it. However, their work does not target realtime animation or dynamics, and the usage of a heavy U-Net for rendering the final result is not possible 
in our setting. Aliev \etal \cite{Aliev19} proposes neural point-based graphics, in which the geometry is represented as a point cloud. Each point is associated with a deep feature, and a neural net computes pixel values based on splatted feature points. While being very flexible in modeling 
various geometric structures, such point-cloud based methods are not yet as efficient as mesh-based representations for modeling dynamic faces, for which the topology is known and fixed.
Yariv \etal \cite{Lior20} models the rendering equation with a neural network that takes the viewing direction, 3D location and surface normals as input. 
Mildenhall \etal \cite{mildenhall2020nerf} proposes a method for synthesizing novel views of complex scenes and models the underlying volumetric scene
with a MLP: the MLP takes a positional encoded 3D coordinate and view direction vector and produces pixel values. 
A closely related idea is presented in \cite{sitzmann2019siren}, where a MLP with sinusoidal activation functions is used to map locations to colors. The spectral properties of mapping smooth, low-dimensional input spaces to high-frequency functions using sinusoidal encodings was further studied in~\cite{tancik2020fourfeat}.
Our method is inspired by these methods in using the Pixel Decoder to render image pixels, but we make innovations to adapt these ideas for the problem
of creating high-quality 3DMFM with lightweight computations, including a learned positional encodings and a dense geometry decoder.
\section{Pixel Codec Avatar}
\label{sec:pca}
\begin{figure*}[t]
    \center
    \includegraphics[width=1\linewidth]{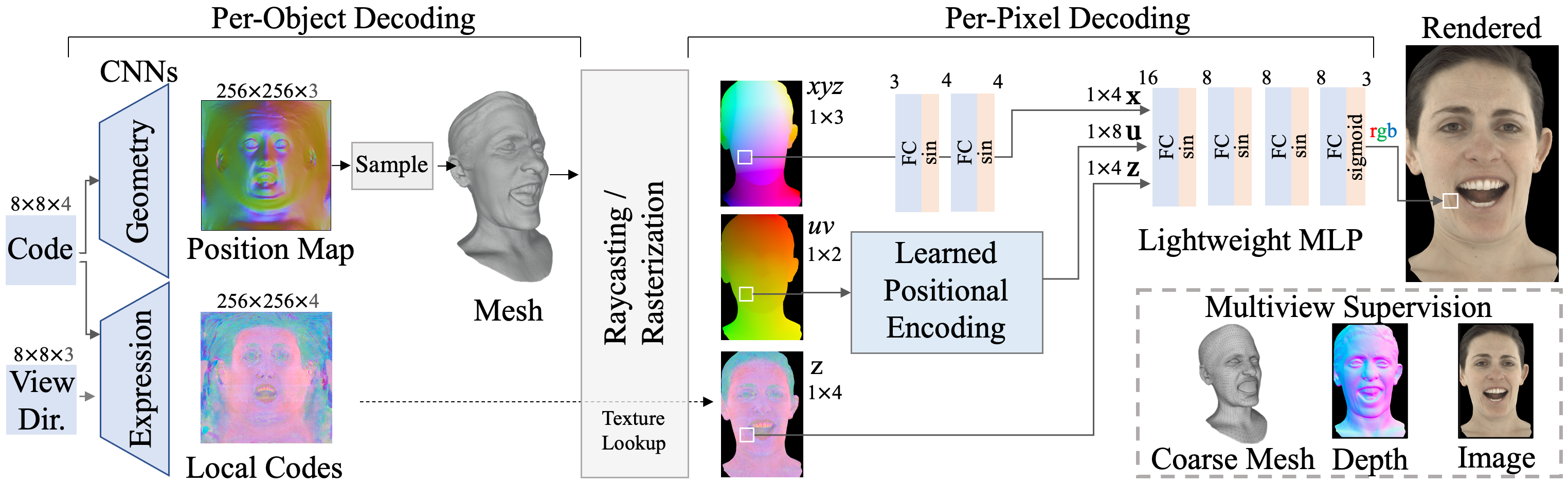}
    \caption{A {\em Pixel Codec Avatar} renders realistic faces by decoding the color of each rasterized or raycast pixel using a shallow SIREN \cite{sitzmann2019siren} that takes as input a local expression code, $\mathbf{z}$, the 3D coordinates in object space, $\mathbf{x}$, and the positional encoded surface coordinates, $\mathbf{u}$, (Section \ref{sec:pe}). This particular combination allows the feature dimensions and network size to remain small and computationally efficient while retaining image fidelity (Section \ref{sec:exp}). The local expression codes and geometry are decoded using fully convolutional architectures from a global latent code and the viewing direction (Section \ref{sec:pca}), and require only small resolutions of $256{\times}256$. Learnable components (in blue) are supervised on multiview images, depth, and tracked coarse mesh.}
   \label{fig:decoder}
\end{figure*}

\begin{figure}[t]
    \center
    \includegraphics[width=\linewidth]{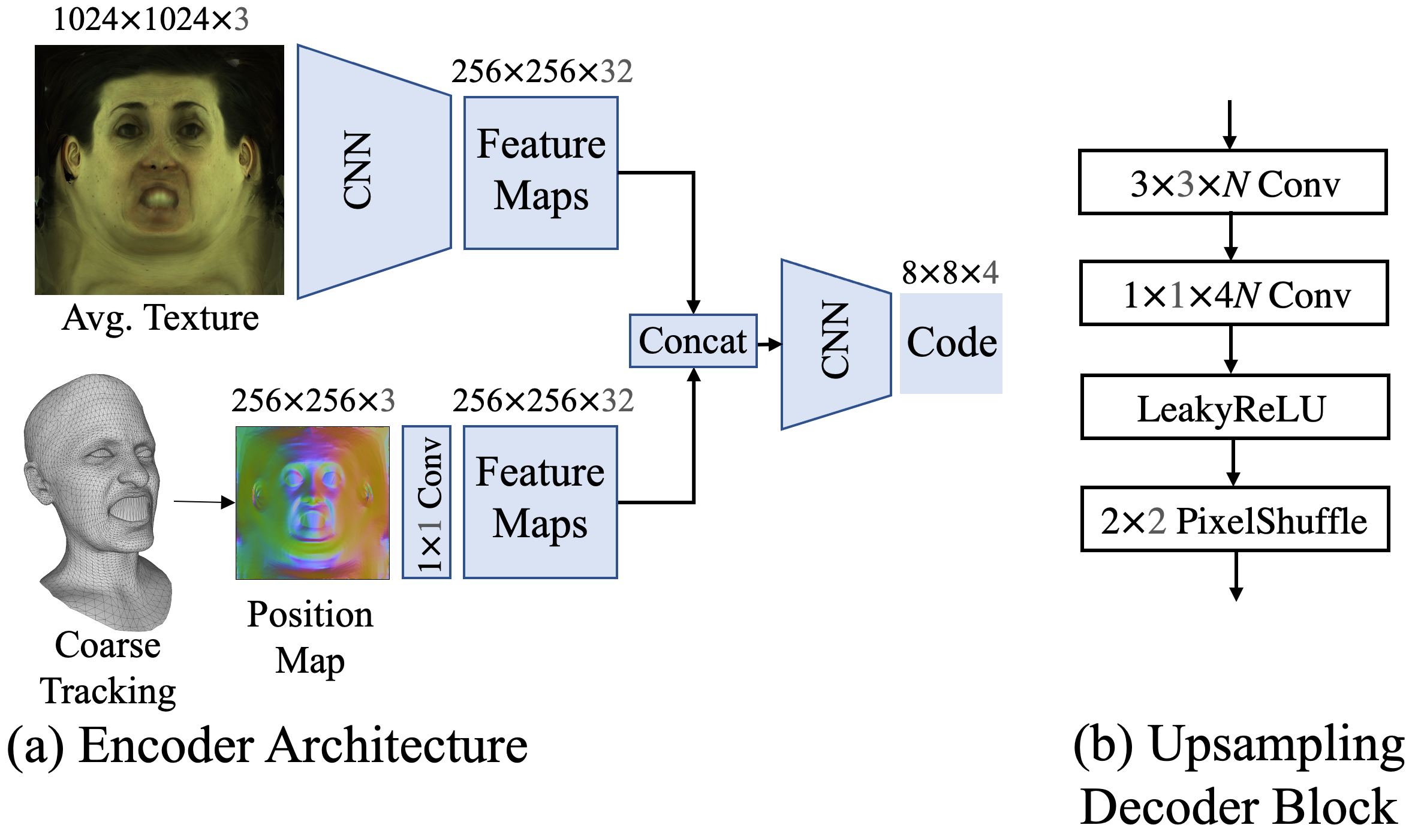}
    \caption{(a) The encoder. (b) The basic block in the geometry decoder and expression decoder.}
    \label{fig:encoder}
    \label{fig:encoder_and_block}
  \label{fig:block}
\end{figure}

The Pixel Codec Avatar is a conditional variational auto-encoder (VAE) where the latent code describes the state of the face (e.g., facial expression) and the decoder produces realistic face images (see Fig.\ref{fig:decoder}) conditioned on a viewing direction. At runtime, latent codes can be produced using a face tracker
to estimate the facial expression (e.g., from cameras mounted on a VR headset~\cite{Stephen18,Shih-En19,Chu2020}), and the estimated code can be used to decode and render realistic face images. At training time, a variational encoder is used to produce the latent codes using multiview training data, similarly to Lombardi \etal ~\cite{Stephen18} (see Fig.~\ref{fig:encoder_and_block}(a)).
The decoder distributes computation across two phases: the {\em Per-Object Decoding} produces the dense mesh and a small map of view conditioned expression codes (Left of Fig.\ref{fig:decoder}), and the {\em Per-Pixel Decoding} computes the on-screen facial pixel values after determining visibility through rasterization or raycasting. 
We use a {\em pixel decoder} $f$ in this second step:
\begin{equation}
    \label{eq:pixdec}
    \mathbf{c} = f(\mathbf{p}), \ \ \ \mathbf{p} = [\mathbf{z}, \mathbf{x}, \mathbf{u}]
\end{equation}
\noindent where $\mathbf{c}$ is the decoded RGB color for a facial pixel, 
and $\mathbf{p}$ is the feature vector for that pixel which is concatenation of the local facial expression code $\mathbf{z}$, the encoded face-centric 3D coordinates $\mathbf{x}$, and the encoded surface coordinates (UV) $\mathbf{u}$. We parameterize $f$ as a small SIREN (see Fig.~\ref{fig:decoder}) and we describe the encoding inputs in Section ~\ref{sec:pe}. The right side of Fig.\ref{fig:decoder} illustrates the Per-Pixel Decoding. We outline the major components:

\vspace{0.5em}
\noindent \textbf{Encoder} (see Fig.~\ref{fig:encoder})(a)) encodes the average texture, computed over unwrapped textures of all camera views, and a tracked mesh into a latent code. Note this tracked mesh is coarse, containing 5K vertices, and doesn't contain vertices for tongue and teeth. We only assume availability of such coarse mesh for training because face tracking using dense mesh over long sequences with explicit teeth and tongue tracking is both challenging and time consuming. Requiring only coarse mesh in training makes our method more practical. In Lombardi \etal \cite{Stephen18}, the 3D coordinates of mesh vertices are encoded using a fully connected layer and fused with texture encoder;
in contrast, we first convert the mesh into a position map using a UV unwrapping of the mesh. Joint encoding of the geometry and texture is then applied, and the final code is a grid of spatial codes, in our case an 8x8 grid of 4 dimensional codes.

\vspace{0.5em}
\noindent \textbf{Geometry Decoder} takes the latent code as input and decodes a dense position map describing face-centric 3D coordinates at each location.
The architecture is fully convolutional, and the basic building block is shown in Fig. \ref{fig:encoder_and_block}(b). We convert the position map to a dense mesh by sampling at each vertex's UV coordinates, and rasterize it to determine visible pixels. In our experiments, the position map is 256$\times$256 and the extracted dense mesh has 65K vertices.

\vspace{0.5em}
\noindent \textbf{Expression Decoder} uses the latent code and the viewing direction to decode a low resolution, view-dependent map of local codes. It consists of the decoder block in Fig. \ref{fig:encoder_and_block}(b) and the output map is 256$\times$256 in our experiments.

\vspace{0.5em}
\noindent \textbf{Pixel Decoder} decodes the color at each facial pixel given $\mathbf{p}$. Specifically, rasterization determines whether a screen pixel corresponds to a visible mesh point, and, if so, the triangle id and barycentric coordinates of the mesh point.
This allows us to compute the encoding inputs $\mathbf{p}$ from the expression map, the vertex coordinates, and the UV coordinates of the triangle.
Inspired by the pixel-wise decoding of images in Sitzmann \etal \cite{sitzmann2019siren}, the {\em pixel decoder} is designed as a SIREN. However, we use a very lightweight network by design, with 4 layers and a total of 307 parameters. We utilize effective encoding in $\mathbf{u}$ to produce facial details with such a light model, described in Section~\ref{sec:pe}.
\section{Positional Encodings for Pixel Decoders}
\label{sec:pe}

While neural networks and MLPs in particular can represent functions of arbitrary complexity when given sufficient capacity, lightweight MLPs tend to produce low-frequency outputs when given smoothly varying inputs~\cite{sitzmann2019siren,tancik2020fourfeat,mildenhall2020nerf}. Thus, given only the smooth face-centric coordinates and surface coordinates as input, a lightweight pixel decoder tends to produce smooth output colors for neighboring pixels, leading to a loss of sharpness in the decoded image. Instead, we encode information about such spatial discontinuities at the input of the Pixel Decoder using two strategies: a low resolution local expression code $\mathbf{z}$ for dynamics, and a learned non-parametric positional encoding $\mathbf{u}$ of surface coordinates for detail. These complement the mesh coordinate input $\mathbf{x}$, which encodes face-centric $xyz$ coordinates using a two-layer SIREN.

\vspace{0.5em}
\noindent \textbf{Facial Expression Positional Encodings} The global expression code, \ie output of the {\it Encoder}, is decoded to a low resolution map of local expression codes (bottom left of Fig.\ref{fig:decoder}) and is further rasterized to the screen space (bottom middle in Fig.\ref{fig:decoder}). This leads to a low dimensional encoding $\mathbf{z}$ of local facial expression at each pixel position. We find it crucial to use the local expression codes for decoding high fidelity facial dynamics.

\vspace{0.5em}
\noindent \textbf{Facial Surface Positional Encodings} 
The local expression codes are too low resolution to capture high-frequency details. We therefore additionally provide the pixel decoder with a positional encoding $\mathbf{u}$ of the facial surface coordinates $(u, v)$ at each pixel. While generic positional encodings such as sinusoids~\cite{mildenhall2020nerf} may achieve highly detailed reconstructions, they require a large number of frequency levels and therefore high dimensionality, incurring computational cost.
Instead, we dramatically reduce the dimensionality of the input features by designing a {\em learned non-parametric positional encoding} function,
\begin{equation}
    \mathbf{u} = [m_{uv}(u, v),\  m_u(u),\  m_v(v)]
    \label{eq:learned_encoding}
\end{equation}
where $m_{uv}$ jointly encodes both $u$ and $v$; $m_u$ and $m_v$ encodes $u$ and $v$ respectively. We directly model $m_{uv}$, $m_{v}$ and $m_{u}$ as non-parametric functions that 
retrieve a low-dimensional encoding from a learned encoding map given $(u, v)$. Specifically, $m_{uv}$ retrives a 4 dimensional vector from a 1024$\times$1024$\times$4 encoding map at position $(u, v)$ using bilinear interpolation; and, similarly,
$m_{u}$ and $m_{v}$ retrieve 2-dimensional vectors from two separate 10000x1 maps respectively. All three maps are jointly learned with the rest of the model. 
Intuitively, $m_{uv}$, $m_u$, and $m_v$ are piece-wise linear functions with 1K$\times$1K breakpoints in 2D, and 10K breakpoints in 1D respectively, and the breakpoints' 
values in the maps contain spatial discontinuity information on the face surface, learned directly from the data. We use 1D encoding
functions $m_{u}$ and $m_{v}$ in addition to the 2D encoding function $m_{uv}$ as a cost-effective way to model higher resolution while avoiding a quadratic increase in model parameters. Empirically, we found that the combination of the two generates better reconstructions than using either one in isolation (Section~\ref{sec:ablation}).

\section{Joint Learning with a Dense Mesh Decoder}
\label{sec:hrgeo}
The geometry used for pixel decoders needs to be accurate and temporally corresponded to prevent the pixel decoders from having to compensate for geometric misalignments via complex view-dependent texture effects.
To achieve this, we learn the variational decoder of geometry and expression jointly with the pixel decoder.

\begin{figure}[t]
    \center
    \includegraphics[width=0.99\linewidth]{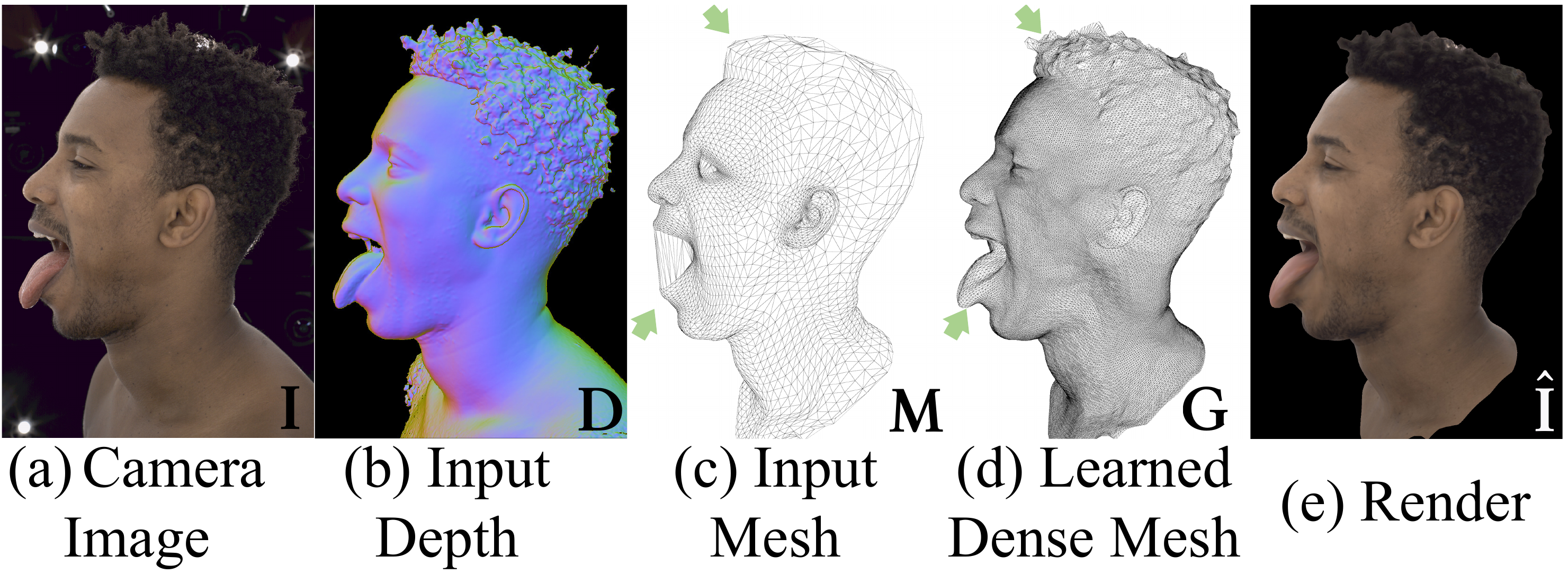}
    \caption{
    We supervise on (a) images, (b) depth, and (c) a coarse tracking mesh of 7K vertices, from which we learn a corresponded, dense face mesh (d) at a higher resolution of ~65K vertices, even in places where the coarse tracked mesh provides no information. The final render (e) can represent difficult-to-track expressions, e.g., involving the tongue.}
   \label{fig:meshes}
\end{figure}

We use a set of multiview images, $\mathbf{I}_t^c$, (i.e., image from camera $c$ at frame $t$), with calibrated intrinsics $\mathbf{K}_c$ and extrinsics, $\mathbf{R}_c|\mathbf{t}_c$. For a subset of frames we compute depth maps $\mathbf{D}^c_t$ using multiview stereo (MVS). 
Additionally, we use a vision-based face tracker to produce a coarse mesh $\mathbf{M}_t$ represented as a position map to provide rough temporal correspondences. 
Note, however, that the input tracked mesh is low resolution, lacking detail in difficult to track areas like the mouth and eyes (Fig.~\ref{fig:meshes}(c)). 
Intuitively, the more accurate the geometry is, the easier and better the pixel decoder may decode the pixel's color. Therefore, our geometry decoder generates a position map $\mathbf{G}$ of a dense mesh of $\sim$65K vertices, including the mouth interior, without direct supervision from a tracked {\em dense} mesh (Fig.~\ref{fig:meshes}(d)).

For each training sample, we compute an average texture $\mathbf{T}_t^{\textrm{avg}}$ by backprojecting the camera images onto the coarse tracking mesh, similarly to~\cite{Stephen18}. 
The texture and the position map computed from the coarse mesh are used as input to the convolutional encoder, $E(\cdot)$, Fig.~\ref{fig:encoder}(a), to produce the latent code $\mathbf{Z} {=} E(\mathbf{T}^{\textrm{avg}}_t, \mathbf{M}_t)\in \mathbb{R}^{8{\times}8{\times}4}$, where the channel dimension is last.
Additionally, we compute the camera viewing direction as $\mathbf{R}_c^T\mathbf{t}_c$ normalized to unit length,
in face-centric coordinates.
We tile this vector into an 8x8 grid $\mathbf{V}{\in}\mathbb{R}^{8{\times}8{\times}3}$.
The geometry and expression decoders in Fig.~\ref{fig:decoder} produce the geometry and local codes,
\begin{align}
\mathbf{G} &= D_g(\mathbf{Z}), & \mathbf{E} &= D_e(\mathbf{Z}, \mathbf{V}),
\end{align}
where $\mathbf{G}{\in}\mathbb{R}^{256{\times}256{\times}3}$ is a position map, and $\mathbf{E}{\in}\mathbb{R}^{256{\times}256{\times}4}$ is a map of expression codes. 
The position map is sampled at each vertex's UV coordinates to produce a mesh for rasterization.
Rasterization assigns to a pixel at screen position $\mathbf{s}$ its corresponding $uv$ coordinates and face-centric $xyz$ coordinates, from which the encoding $\mathbf{p}$ is derived as described in Sect.~\ref{sec:pe}. 
The final pixel color is decoded producing a rendered image, $\hat{\mathbf{I}_t^c}(\mathbf{s}) {=} f(\mathbf{p})$.
At each SGD step, we compute a loss
\begin{equation}
\label{eq:losses}
\mathcal{L} = \lambda_i\mathcal{L}_{I} + \lambda_d\mathcal{L}_{D} + \lambda_n\mathcal{L}_{N} + \lambda_m\mathcal{L}_{M} + \lambda_s\mathcal{L}_{S}+ \lambda_{kl}\mathcal{L}_{KL}\;,
\end{equation}
\noindent where $\mathcal{L}_{I}{=}||\mathbf{I}_t^c{-}\hat{\mathbf{I}_t^c}||_2$ measures image error, and $\mathcal{L}_{D}{=}||(\mathbf{D}_t^c{-}\hat{\mathbf{D}_t^c})\odot \mathbf{W}_D||_1$ measures depth error, where $\mathbf{W}_D$ is a mask 
selecting regions where the depth error is below a threshold of 10mm.
We additionally use a normal loss, $\mathcal{L}_{N}{=}||(N(\mathbf{D}_t^c){-}N(\hat{\mathbf{D}_t^c})){\odot} \mathbf{W}_D||_2$ where $N(\cdot)$ computes normals in screen space and encourages sharper geometric details. 
The remaining terms are regularizations: $\mathcal{L}_{M}{=}||(S(\mathbf{G}){-}S(\mathbf{M}_t)){\odot} \mathbf{W}_M||_2$, where $S(\cdot)$ is a function that samples the position map at the vertex UVs, penalizes large deviations from the coarse tracking mesh using a mask $\mathbf{W}_M$ to avoid penalizing the mouth area (where the tracked mesh is inaccurate). $\mathcal{L}_{S}$ is a Laplacian smoothness term~\cite{LaplacianMeshEditing:2004} on the dense reconstructed mesh. 
These terms prevent artifacts in the geometry stemming from noise in the depth reconstructions, images with no depth supervision, and noisy SGD steps.
Implementation details for the smoothness term and on how differentiable rendering is used to optimize these losses can be found in the supplemental materials.
$\mathcal{L}_{KL}$ is the Kullback-Leibler divergence term of the variational encoder.

The above procedure recovers detailed geometry in the decoded dense mesh that is not captured in the input tracked meshes.
Especially note-worthy is the automatic assignment of vertices inside the mouth to the teeth and tongue, as well as hair, see Fig. \ref{fig:qualitative} for examples. 

\section{Experiments}
\label{sec:exp}

\begin{table}
    \begin{center}
    {\footnotesize
    \begin{tabular}{l|c|c|c|c|c|c}
    \hline
     & \bf{Model} & \bf{Front} &\bf{Up} &\bf{Down} &\bf{Left} &\bf{Right} \\
    \hline \hline
     \multirow{4}{*}{S1} & Baseline    & 23.03 & 20.78 & 18.13 & 16.32 & 18.97\\
                         & \bf{Full}   & \bf{21.39} & \bf{19.71} & \bf{17.52} & \bf{15.52} & \bf{18.00}\\
                         & No-UV        & 22.16 & 20.38 & 18.28 & 16.27 & 18.57\\
                         & Coarse      & 21.64 & 20.04 & 17.84 & 16.02 & 18.69\\
    \hline
     \multirow{4}{*}{S2} & Baseline    & 19.53 & 20.90 & 16.62 & 15.44 & 13.52\\
                         & \bf{Full}   & \bf{18.31} & \bf{19.96} & \bf{16.36} & \bf{14.28} & \bf{12.14}\\
                         & No-UV        & 19.34 & 20.52 & 17.61 & 15.40 & 13.29\\
                         & Coarse      & 19.88 & 21.62 & 17.97 & 15.97 & 13.92\\
     \hline
     \multirow{4}{*}{S3} & Baseline    & 24.41 & 22.83 & 16.54 & 16.09 & 16.81\\
                         & \bf{Full}   & \bf{23.11} & \bf{22.22} & \bf{16.04} & \bf{15.29} & \bf{15.64}\\
                         & No-UV        & 23.95 & 22.99 & 16.42 & 15.86 & 16.12\\
                         & Coarse      & 23.94 & 23.04 & 16.44 & 15.81 & 16.79\\
    \hline
     \multirow{4}{*}{S4} & Baseline    & 7.26 & 6.03 & 7.34 & 7.15 & 7.76\\
                         & \bf{Full}   & \bf{6.81} & \bf{5.78} & \bf{7.33} & \bf{7.05} & \bf{7.63}\\
                         & No-UV        & 7.20 & 6.13 & 7.40 & 7.32 & 8.05\\
                         & Coarse      & 7.19 & 6.02 & 7.48 & 7.21 & 8.25\\
    \hline
     \multirow{4}{*}{S5} & Baseline    & 9.20 & 10.87 & 7.24 & 7.27 & 6.54\\
                         & \bf{Full}   & \bf{8.74} & \bf{10.37} & \bf{7.16} & \bf{7.09} & \bf{6.53}\\
                         & No-UV        & 9.06 & 10.96 & 7.39 & 7.46 & 6.76\\
                         & Coarse      & 9.09 & 10.64 & 7.49 & 7.49 & 6.56\\
     \hline
     \multirow{4}{*}{S6} & Baseline    & 6.86 & 6.53 & 5.85 & 5.66 & 5.29\\
                         & \bf{Full}   & \bf{6.22} & \bf{6.06} & \bf{5.39} & \bf{4.97} & \bf{4.95}\\
                         & No-UV        & 6.86 & 6.72 & 5.85 & 5.90 & 5.62\\
                         & Coarse      & 6.54 & 6.33 & 5.69 & 5.29 & 5.16\\
    \hline\hline
    \end{tabular}
    }
    \end{center}
    \caption{\textbf{MSE} on pixel values of the rendered images against the ground truth images on test set, evaluated on 5 views. {\em Baseline} is the model in \cite{Stephen18}; {\em Full} is our model PiCA (Fig.\ref{fig:decoder}), {\em No-UV} is PiCA variant that is not using surface coordinates; {\em Coarse} is PiCA variant that decodes coarse mesh (7K vertices). {\em Full} PiCA model consistently outperform others on all tested identities over all views.}
    \label{tab:results}
\end{table}

\begin{table}
    \begin{center}
    {\footnotesize
    \begin{tabular}{l|c|c|c|c|c|c}
    \hline
     & \bf{Model} & \bf{Front} &\bf{Up} &\bf{Down} &\bf{Left} &\bf{Right} \\
    \hline \hline
      
    \multirow{5}{*}{S1} & \bf{Full}   & \bf{21.39} & 19.71 & \bf{17.52} & \bf{15.52} & \bf{18.00}\\
                         & NERF-PE     & 21.85 & 20.10 & 17.86 & 15.90 & 18.61\\
                         & UV-NoPE     & 21.45 & 19.93 & 17.70 & 15.98 & 18.53\\
                         & 2D-PE       & 21.56 & 19.85 & 17.97 & 15.98 & 18.80\\
                         & 1D-PE       & 21.40 & \bf{19.67} & 17.60 & 15.70 & 18.29\\
                         
    \hline
     \multirow{5}{*}{S2} 
                         & \bf{Full}   & \bf{18.31} & \bf{19.96} & \bf{16.36} & \bf{14.28} & \bf{12.14}\\
                         & NERF-PE     & 18.99 & 20.35 & 17.35 & 15.19 & 13.18\\
                         & UV-NoPE     & 19.17 & 20.51 & 17.53 & 15.40 & 13.29\\
                         & 2D-PE       & 19.05 & 20.23 & 17.47 & 15.02 & 13.02\\
                         & 1D-PE       & 19.30 & 20.61 & 17.64 & 15.43 & 13.39\\
    \hline
     \multirow{5}{*}{S6} 
                         & \bf{Full}   & \bf{6.22} & 6.06 & \bf{5.39} & \bf{4.97} & \bf{4.95}\\
                         & NERF-PE     & 6.41 & 6.16 & 5.60 & 5.29 & 5.14\\
                         & UV-NoPE     & 6.59 & 6.53 & 5.68 & 5.33 & 5.24\\
                         & 2D-PE       & 6.28 & \bf{6.00} & 5.48 & 5.26 & 5.09\\
                         & 1D-PE       & 6.58 & 6.39 & 5.68 & 5.26 & 5.21\\
    \hline\hline
    \end{tabular}
    }
    \end{center}
    \caption{Ablation on usage of UV coordinates: encoding with learned encoding maps ({\em Full}), directly using UV ({\em UV-NoPE}), encoding with sinusoidal functions \cite{mildenhall2020nerf} ({\em NERF-PE} ), joint encoding only ({\em 2D-PE}) and separate encoding only ({\em 1D-PE})}
    \label{tab:uv_encoding}
\end{table}

\begin{table}
    \begin{center}
    {\scriptsize
    \begin{tabular}{l|c|c|c|c}
    \hline \hline
    & & {\bf 18cm (2.7M)} & {\bf 65cm (0.9M)} & {\bf 120cm (0.2M)} \\ \hline
    \hline
    \multirow{2}{*}{DSP Step} & Baseline & 44.76 ms & 44.76 ms & 44.76 ms \\ \cline{2-5}
                              & PiCA & 2.16 ms & 2.16 ms & 2.16ms\\ \cline{2-5}                     
    \hline \hline
    \multirow{2}{*}{GPU Step} & Baseline & 2.67 ms & 2.47 ms & 1.94 ms\\ \cline{2-5}
                         & PiCA & 8.70 ms & 3.27 ms & 2.70 ms\\ 
    \hline \hline
    \end{tabular}
    }
    \end{center}
    \caption{Runtime performance on the Oculus Quest 2, measured at 3 different avatar distances (the numbers in parenthesis are avatar pixels to render). Note that 60-120cm are typical interpersonal distances \cite{social_distance}, while 18cm would be considered intimate. }
    \label{tab:runtime}
\end{table}

\begin{figure}[t]
    \center
    \includegraphics[width=1\linewidth]{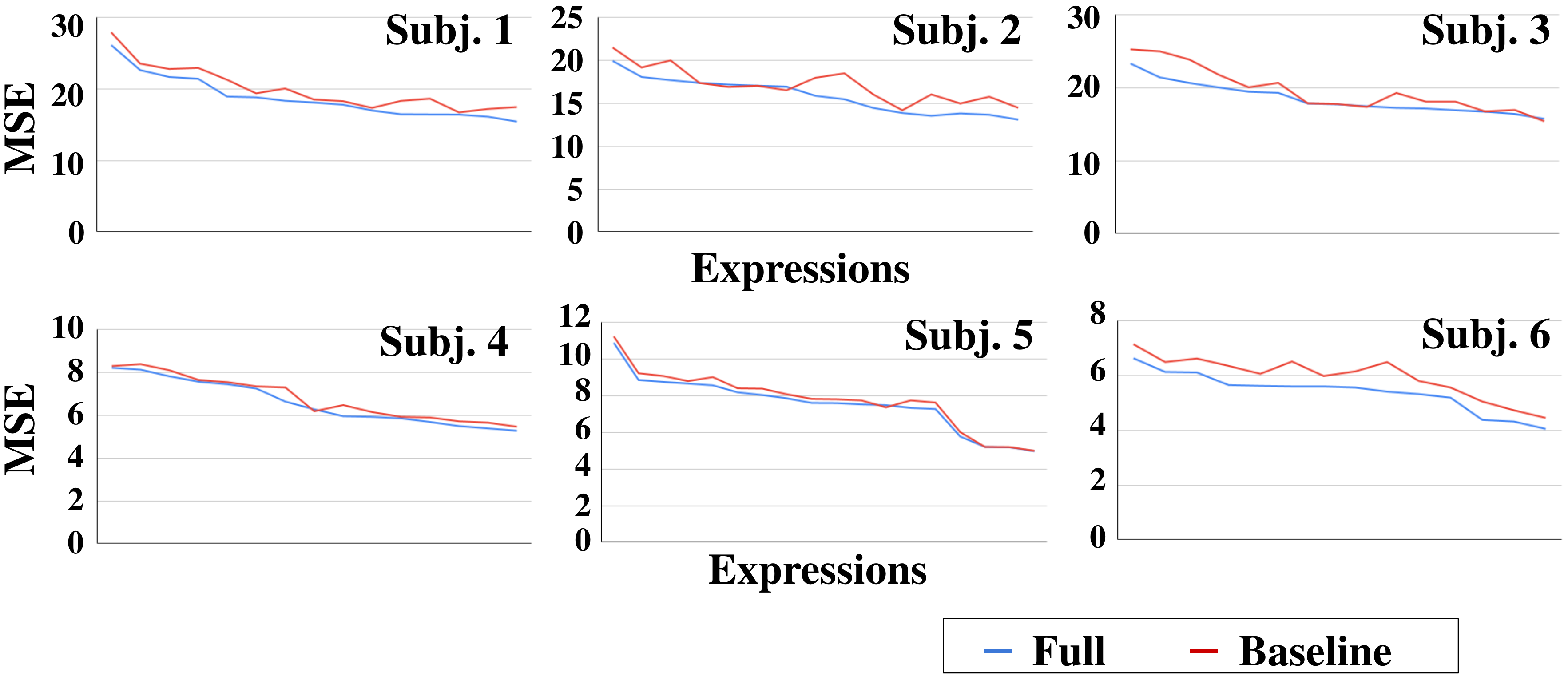}
    \caption{The \textbf{MSE} distribution over test expressions, sorted in decreasing order for the {\em Full} model: x-axis is expressions and y-axis is MSE. We can see that the performance of our model is similar or better than the baseline across expressions for all identities.}
   \label{fig:err_distr}
\end{figure}

\begin{figure*}[t]
    \center
    \includegraphics[width=1\linewidth]{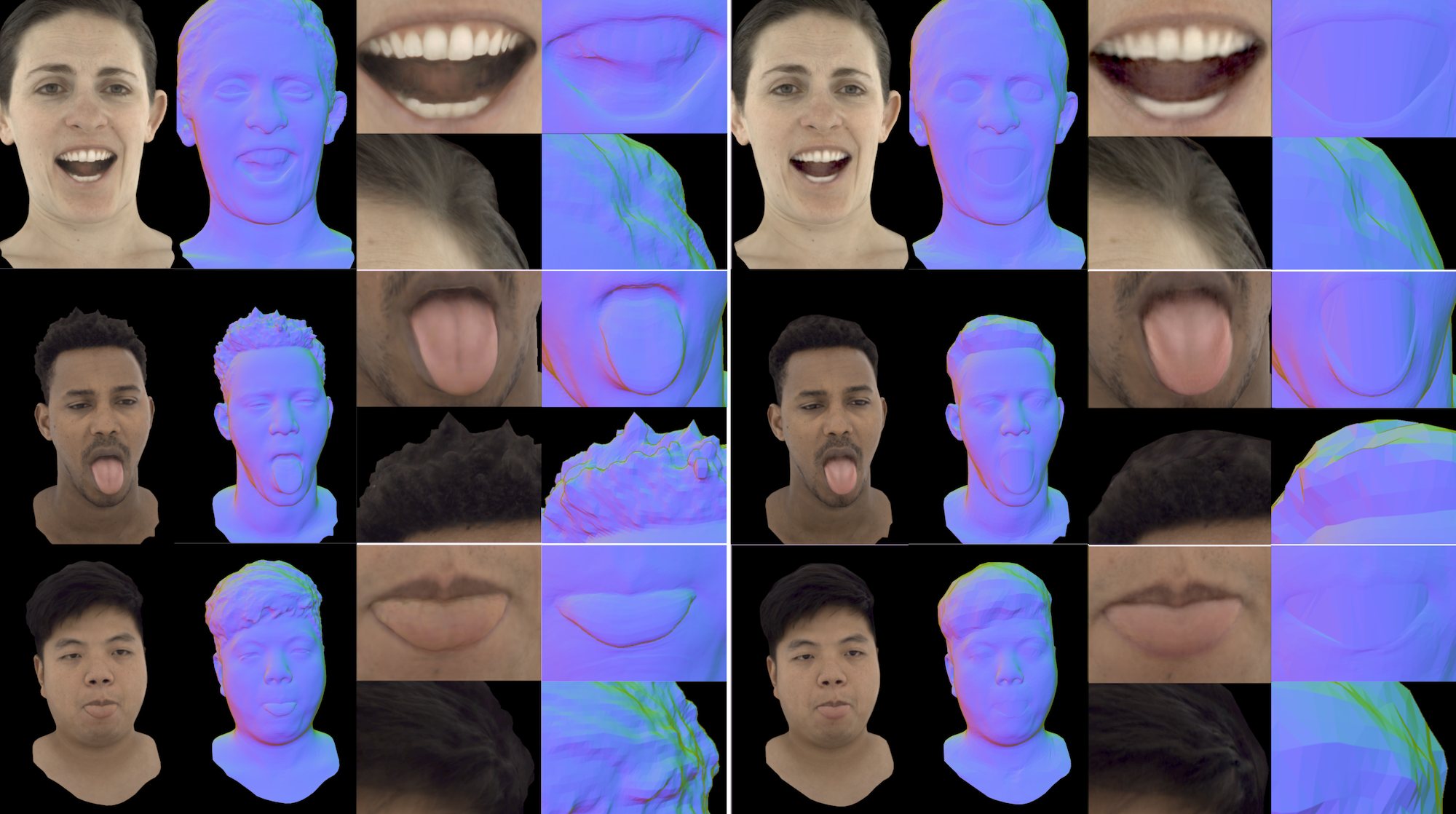}
    \caption{Example rendered faces comparing our {\em Full} model (left) with the baseline \cite{Stephen18} (right). For each example, we show the rendered full face and the depth image, and close looks for two facial regions. The visual qualities of rendered images are good for both models, while our model produce sharper details at teeth, tongue and hair. The depth images show more geometry details generated by our model.}
    \label{fig:qualitative}
\end{figure*}

\vspace{0.5em}
\noindent \textbf{Experiment Setting} We evaluate our model on 6 identities on 5 different viewing directions: front, upward, downward, left and right (see example images in the supplemental material). We capture multiview video data for each identity using two face capture systems: Subject 1-4 are captured with 40 cameras with 50mm focal length, while Subject 5 and 6 are captured with 56 cameras at 35mm focal length.
We use images of size 2048$\times$1334 for training and testing.
The data of each identity consists of expressions, range of facial motion, and reading sentences. We randomly select expressions and sentence readings as testing data, leading to $\sim$12K frames for training and $\sim$1K frames for testing per identity. The total number of images is roughly the number of frames multiplied by the number of cameras. All models are trained with batchsize 4, at learning rate 0.001, for 400000 iterations. The weights for different loss terms in Eq.~\ref{eq:losses} for $\lambda_i$, $\lambda_d$, $\lambda_n$, $\lambda_m$, $\lambda_s$ and $\lambda_{kl}$ are set to 2, 10, 1, 0.1, 1 and 0.001 respectively. We report Mean Squared Error (MSE) between rendered image and original image on rasterized pixels on testing data as the evaluation metric for reconstruction quality. Note that the results of different identities are not directly comparable due to different camera settings and subject appearance such as facial skin tone and hair style.

\subsection{Overall Performance}
The baseline model has 19.08{\it M} parameters and PiCA has 5.47{\it M}. In particular, the pixel decoder of PiCA only has 307 parameters. {\bf When rendering 5 avatars (evenly spaced in a line, 25cm between neighboring pair) in the same scene on a Oculus Quest 2, PiCA runs at $\sim$50 FPS on average, showing the possibility of multi-way telepresence call}. 
\textbf{In Table \ref{tab:results} and Fig.~\ref{fig:err_distr} we report quantitative comparisons which show PiCA consistently achieves better reconstruction across all tested identities, expressions and views, despite a 3.5$\times$ reduction in model size and much faster computation (Table~\ref{tab:runtime}).} Specifically, Table~\ref{tab:results} compares the reconstruction quality over 5 views, averaged over all testing expressions. Fig.~\ref{fig:err_distr} plots MSE values of Full and Baseline over all testing expressions (sorted in decreasing order of Full's results). 
\begin{figure}[t]
    \center
    \includegraphics[width=1\linewidth]{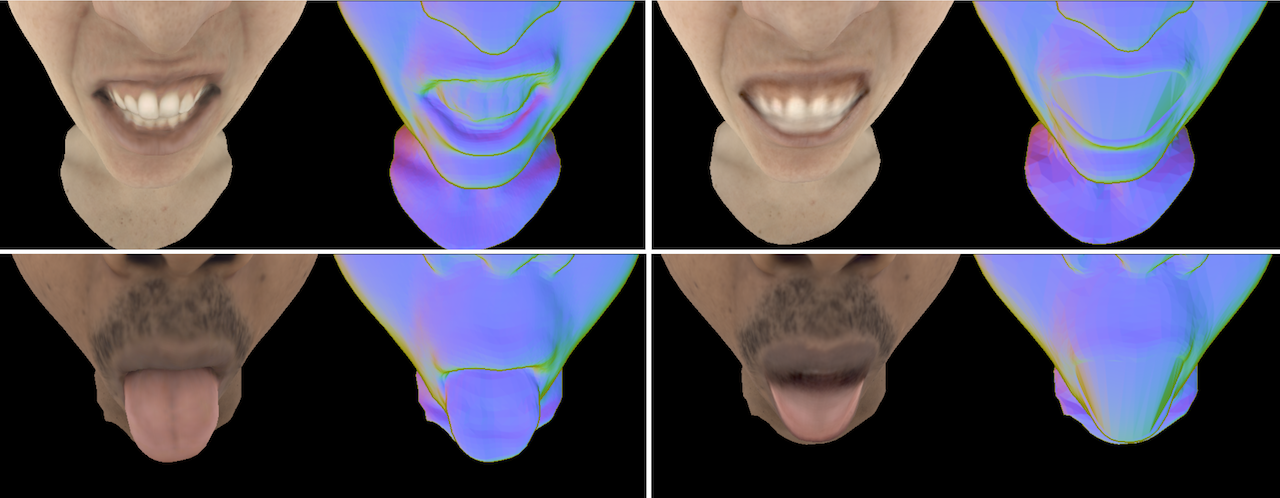}
    \caption{Rendering at a novel viewing position, much closer to the mouth than any training views. Two example frames are shown with the rendered depth as well: left column is PiCA Full, and right is the Baseline model \cite{Stephen18}, best viewed when magnified.}
   \label{fig:nearfield}
   \vspace{-1.5em}
\end{figure}
Qualitative examples are shown in Fig.~\ref{fig:qualitative} and we invite the readers to see more high resolution results in supplemental materials. Example result frames for both our Full model (left) and the baseline model (right) are shown, and we also show local regions at higher resolution for closer inspection. Overall, both models produce very realistic looking faces. Our model produces sharper results in many facial regions, especially the selected regions showing teeth, tongue, and hair. 

\subsection{Ablation Studies}
\label{sec:ablation}
\vspace{0.3em}
\noindent \textbf{UV Positional Encoding} 
Many details of the facial surface is represented as discontinuities in color values in neighboring pixels, \eg a skin pixel adjacent to a hair pixel. We model such discontinuities with learned encoding maps such that the encoding function is piece-wise linear with the map entries as the learned breakpoint values (Section~\ref{sec:pe}). In this section, we study the benefit of this proposed method. 
We train a PiCA variant {\em No-UV} that does not use UV coordinates for decoding pixel values. In Table~\ref{tab:results} one can see that Full PiCA model consistently outperforms the No-UV variant, showing clear advantage of using encoded UV coordinates. We Further compare our approach with directly using UV without encoding, and encoding UV with sinusoidal functions \cite{mildenhall2020nerf}. We train two additional PiCA variants {\em UV-NoPE} that uses UV without any encoding, and {\em NERF-PE} that encodes UV using the encoding function of \cite{mildenhall2020nerf} (a 40-dimensions code compared to 8-dimensions for Eq.~\eqref{eq:learned_encoding}). The comparison results are shown in Table~\ref{tab:uv_encoding}. The {\em Full} model consistently outperforms both variants over all tested views and subjects, proving the effectiveness of encoding UV with learned encoding maps. We also ablate on our encoding scheme: we train a PiCA variant {\em 2D-PE} that only performs 2D joint encoding ($m_{uv}$ in Eq.~\eqref{eq:learned_encoding}) and {\em 1D-PE} that only performs 1D separate encodings  ($m_{u}, m_{v}$). The comparison results are shown in Table~\ref{tab:uv_encoding}. The Full PiCA model combining both joint encoding and 1D encodings outperforms these two variants, showing that the two encoding methods are complementary and by combining both we can achieve consistent performance improvement.

\vspace{0.3em}
\noindent \textbf{Dense Mesh Decoder}
In Fig.~\ref{fig:qualitative}, we show depth images alongside the rendered images. The dense mesh generated by our model contains more geometry information and the corresponding rendered images are sharper: in particular, one may inspect the teeth, tongue and hair regions. In Fig.~\ref{fig:nearfield} we compare novel viewpoint rendering results of Full and Baseline at a viewing position that is very close to the mouth: there are no such views in our training set. While the baseline results look like a pasted plane inside the mouth, ours look more realistic thanks to the more accurate geometry in the generated dense mesh \eg at teeth, tongue and lips.  
For quantitative study, we train a PiCA model variant {\em Coarse} which decodes coarse meshes of the same topology used in \cite{Stephen18}. In Table~\ref{tab:results}, we evaluate it on the test set, and the results show it being consistently inferior to the Full PiCA model, illustrating the benefit of the dense geometry decoder in the Pixel Codec Avatar. 

\subsection{Runtime Performance on Mobile SoC}
\label{sec:runtime}
We present runtime performance on a Oculus Quest 2 VR headset  \footnote{The baseline model and the geometry and expression decoders of PiCA are 8-bit quantized to execute on the DSP, with small quality drops.} in Table \ref{tab:runtime}. We measure the time spent on both the DSP (Digital-Signal-Processing unit) and the GPU steps - note the two steps are pipelined at runtime. There is 20$\times$ reduction for DSP time from Baseline to PiCA. Overall, Baseline runs at $\sim$22 FPS, while PiCA hits the Quest 2's maximum framerate at 90 FPS. While the baseline model always decodes entire texture map of the avatar head at fixed resolution, PiCA decodes only visible regions with resolution adaptive to the distance of the avatar. Further more, PiCA allows a pipelined decoding process balanced in computation load distribution on a mobile SoC: while the per-object decoding needs to be done on the DSP for the convolution operations, the lightweight pixel decoder can be implemented in the highly optimized fragment shader so that the per-pixel decoding can be done on the GPU. In contrast, for the baseline model the decoding computation of the mesh and the texture needs to be done entirely on the DSP and the GPU only performs the final rendering given decoded texture and mesh.

\section{Conclusion and Future Work}
\label{sec:conclusion}

We present the Pixel Codec Avatar as a high quality lightweight deep deformable face model, as a potential technology for enabling multi-person telecommunication in virtual reality on a mobile VR headset. This work only focuses on the {\em decoder} and we can follow the method in Wei \etal \cite{Shih-En19} to build the {\em encoder} for the telepresence communication system. Achieving high fidelity low latency telepresence communication by improving the encoder and decoder models is the main direction for future work. 
\appendix
\section{Appendices}
\subsection{Encoder and Decoder Architectures}
\noindent \textbf{Encoder} The encoder consists of three major components: the {\em tex-head}, {\em geom-head} and the {\em tex-geom-encoder}. The {\em tex-head} has two blocks of {\em conv+leakyrelu}, where the {\em conv} for both layers have kernel size 4, stride 2. The first one has 512 output channels, and the second has 256 channels. The {\em geom-head} has one block of {\em conv+leakyrelu} where the {\em conv} has kernel size 1, stride 1 and output channel number 256. The output of {\em tex-head} and {\em geom-head} are both 256x256x256, and are concatenated and passed to {\em tex-geom-encoder}, which has 5 blocks of {\em conv+leakyrelu}. The kernel size and stride are all 4 and 2 for all {\em conv}, while the output channel numbers are 128, 64, 32, 16 and 8 respectively. The output of {\em tex-geom-encoder} is further passed to two separate 1x1 {\em conv} layers to produce mean and variance. {\em leakyrelu} always having leaky threshold set to 0.2.

\vspace{0.5em}
\noindent \textbf{Per-Object Decoder} This decoder decodes the local expression code and the dense mesh from the latent code which is of dimension 8x8x4. It consists the {\em geometry decoder}, containing 5 blocks of the building block showing in Fig. 3b in the main text, with output channel numbers 32, 16, 16, 8, 3 respectively. The output size is 256x256x3, from which the dense mesh can be retrieved using the uv coordinates of the mesh vertices. The {\em expression decoder} takes the concatenated latent code and view direction as input, which is of size 8x8x7, and it contains 5 building blocks as showing in Fig. 3b as well, with output channel numbers 32, 16, 16, 8, 4 respectively. Note in both cases the first {\em conv} in the block in Fig. 3b has a per-channel per-spatial location bias parameter, following \cite{Stephen18}.

\vspace{0.5em}
\noindent \textbf{Pixel Decoder} The entries in the 2D and 1D encoding maps in the pixel decoder are initialized to have uniform distribution in the range [-1, 1]. The 3D coordinate input (x,y,z) are first converted to a 4-dimensional vector via a two layer SIREN with output channel numbers 4 and 4 respectively, and then it is concatenated with the encoded uv (8-dimension) and the local expression code to form a 16 dimensional input to the final SIREN. The final SIREN has 4 layers with output channel numbers 8, 8, 8, 3 respectively to compute the RGB color at a pixel.     

\subsection{Geometric Smoothness}
To recap, $\mathbf{G}\in\mathbb{R}^{w{\times}w{\times}3}$, with $w{=}256$, is a decoded position map describing the geometry, and $S(\cdot):\mathbb{R}^{w{\times}w{\times}3} \rightarrow \mathbb{R}^{N_V{\times}3}$ is a function that bilinearly interpolates the position map at the vertex $uv$ locations to produce face-centric $xyz$ locations for the set of $N_{V}$ mesh vertices, where $N_{V}$ is the number of vertices in a fixed mesh topology.
Our geometric smoothness regularization term $\mathcal{L}_{S}$ combines two common gradient-based smoothness energies,
\begin{align}
\mathcal{L}_{S} &= \lambda_g \left[ ||\mathcal{D}_x(\mathbf{G}) ||_2 + ||\mathcal{D}_y(\mathbf{G}) ||_2 \right]\\
&+ \lambda_{l}||\mathbf{W}_L\mathbf{L} (S( \mathbf{G} ) - \mathbf{V}_\mu)||_2,
\end{align}
where we identify:

\vspace{0.5em}
\noindent \textbf{Gradient Smoothness}. 
Linear operators $\mathcal{D}_*$ compute the $x$ and $y$ derivatives of the position map using finite differences. 
These terms prevent large changes across neighboring texels in the position map itself. 

\noindent \textbf{Mesh Laplacian}. %
The linear operator $\mathbf{L}\in\mathbb{R}^{N_V{\times}N_V}$ represents the mesh Laplacian discretized using cotangent weights~\cite{LaplacianMeshEditing:2004} computed on the coarse neutral input mesh. 
Here, $\mathbf{V}_\mu\in\mathbb{R}^{N_V{\times}3}$ is a mean face mesh used as a regularization target. 
The diagonal matrix $\mathbf{W}_L\in\mathbb{R}^{N_V{\times}N_V}$ weights the regularization on hair and mouth vertices at 1.25 and the remaining vertices at 0.25.
This regularization prevents the differential mesh coordinates (as computed by the mesh Laplacian) from deviating excessively from the regularization target. 

The regularization target $\mathbf{V}_\mu$ is initialized with the coarse neutral mesh geometry. However, because the coarse geometry lacks detail in the mouth, hair, and eye regions, using it as a regularization target tends to oversmooth these areas. Therefore, we update the target on the fly during training using exponential smoothing, obtaining a slowly-changing, moving average estimate of the mean face geometry at dense resolutions. At every SGD iteration, we update $\mathbf{V}_\mu$ as follows:
\begin{equation}
    \mathbf{V}_\mu \leftarrow (1-\lambda_\mu)\mathbf{V}_\mu + \lambda_\mu \frac{1}{B} \sum_{b=1}^B S(\mathbf{G}_b),
    \label{eq:mean_mesh_update}
\end{equation}
where $\lambda_\mu=1e^{-4}$ and $b\in\{1{\ldots}B\}$ iterates over samples in the SGD batch. No SGD gradients are propagated by the update in Eq.~\eqref{eq:mean_mesh_update}. 

In our experiments, we set $\lambda_l=0.1$ and $\lambda_g=1$.
\subsection{Differentiable Rasterizer}
We use a differentiable rasterizer for computing the screen space inputs given dense mesh and local expression code map, as illustrated in Fig.2 in the main text. Note that the geometry information affects the final decoded image via two gradient paths: one is in the rasterization, and the other is as input to the pixel decoder. We empirically found that allowing gradient from the image loss to pass to the geometry decoder from both paths leads to unstable training and geometry artifacts, so we disable the second gradient path mentioned above to achieve stable training. Intuitively, this is to enforce that the geometry decoder should focus on producing correct facial shape, instead of {\em coordinating} with the pixel decoder to produce correct color values.  
{\small
\bibliographystyle{ieee_fullname}
\bibliography{main}

\begin{thebibliography}{10}\itemsep=-1pt

\bibitem{Abrevaya_2019_ICCV}
Victoria~Fernandez Abrevaya, Adnane Boukhayma, Stefanie Wuhrer, and Edmond
  Boyer.
\newblock A decoupled 3d facial shape model by adversarial training.
\newblock October 2019.

\bibitem{Aliev19}
Kara-Ali Aliev, Artem Sevastopolsky, Maria Kolos, Dmitry Ulyanov, and Victor
  Lempitsky.
\newblock Neural point-based graphics.
\newblock {\em arXiv preprint arXiv:1906.08240}, 2019.

\bibitem{Bagautdinov_2018_CVPR}
Timur Bagautdinov, Chenglei Wu, Jason Saragih, Pascal Fua, and Yaser Sheikh.
\newblock Modeling facial geometry using compositional vaes.
\newblock June 2018.

\bibitem{Blanz99}
Volker Blanz and Thomas Vetter.
\newblock A morphable model for the synthesis of 3d faces.
\newblock page 187–194, 1999.

\bibitem{cheng2019meshgan}
Shiyang Cheng, Michael Bronstein, Yuxiang Zhou, Irene Kotsia, Maja Pantic, and
  Stefanos Zafeiriou.
\newblock Meshgan: Non-linear 3d morphable models of faces, 2019.

\bibitem{Chu2020}
Hang Chu, Shugao Ma, Fernando~De la Torre, Sanjia Fidler, and Yaser Sheikh.
\newblock Expressive telepresence via modular codec avatars.
\newblock 2020.

\bibitem{GAN}
Ian Goodfellow, Jean Pouget-Abadie, Mehdi Mirza, Bing Xu, David Warde-Farley,
  Sherjil Ozair, Aaron Courville, and Yoshua Bengio.
\newblock Generative adversarial nets.
\newblock 27:2672--2680, 2014.

\bibitem{VAE}
Diederik~P Kingma and Max Welling.
\newblock Auto-encoding variational bayes.
\newblock 2014.

\bibitem{Lewis2014PracticeAT}
J.~P. Lewis, K. Anjyo, T. Rhee, M. Zhang, F. Pighin, and Zhigang Deng.
\newblock Practice and theory of blendshape facial models.
\newblock 2014.

\bibitem{Stephen18}
Stephen Lombardi, Jason Saragih, Tomas Simon, and Yaser Sheikh.
\newblock Deep appearance models for face rendering.
\newblock {\em TOG}, 37(4), 2018.

\bibitem{mildenhall2020nerf}
Ben Mildenhall, Pratul~P. Srinivasan, Matthew Tancik, Jonathan~T. Barron, Ravi
  Ramamoorthi, and Ren Ng.
\newblock Nerf: Representing scenes as neural radiance fields for view
  synthesis.
\newblock 2020.

\bibitem{Raitt04}
Bay Raitt.
\newblock The making of gollum.
\newblock Presentation at U. Southern California Institute for Creative
  Technologies’s Frontiers of Facial Animation Workshop, August 2004.

\bibitem{Ranjan_2018_ECCV}
Anurag Ranjan, Timo Bolkart, Soubhik Sanyal, and Michael~J. Black.
\newblock Generating 3d faces using convolutional mesh autoencoders.
\newblock September 2018.

\bibitem{richard2020audio}
Alexander Richard, Colin Lea, Shugao Ma, Juergen Gall, Fernando de~la Torre,
  and Yaser Sheikh.
\newblock Audio- and gaze-driven facial animation of codec avatars.
\newblock 2021.

\bibitem{Gabriel20}
Gabriel Schwartz, Shih-En Wei, Te-Li Wang, Stephen Lombardi, Tomas Simon, Jason
  Saragih, and Yaser Sheikh.
\newblock The eyes have it: An integrated eye and face model for photorealistic
  facial animation.
\newblock {\em ACM Trans. Graph.}, 39(4), 2020.

\bibitem{Shamai2019}
Gil Shamai, Ron Slossberg, and Ron Kimmel.
\newblock Synthesizing facial photometries and corresponding geometries using
  generative adversarial networks.
\newblock {\em ACM Trans. Multimedia Comput. Commun. Appl.}, 15(3s), 2019.

\bibitem{sitzmann2019siren}
Vincent Sitzmann, Julien~N.P. Martel, Alexander~W. Bergman, David~B. Lindell,
  and Gordon Wetzstein.
\newblock Implicit neural representations with periodic activation functions.
\newblock 2020.

\bibitem{Slossberg_2018_ECCV_Workshops}
Ron Slossberg, Gil Shamai, and Ron Kimmel.
\newblock High quality facial surface and texture synthesis via generative
  adversarial networks.
\newblock September 2018.

\bibitem{LaplacianMeshEditing:2004}
Olga Sorkine, Daniel Cohen-Or, Yaron Lipman, Marc Alexa, Christian R\"{o}ssl,
  and Hans-Peter Seidel.
\newblock Laplacian surface editing.
\newblock pages 179--188, 2004.

\bibitem{social_distance}
Agnieszka Sorokowska, Piotr Sorokowski, Peter Hilpert, Katarzyna Cantarero,
  Tomasz Frackowiak, Khodabakhsh Ahmadi, Ahmad~M. Alghraibeh, Richmond
  Aryeetey, Anna Bertoni, Karim Bettache, Sheyla Blumen, Marta Błażejewska,
  Tiago Bortolini, Marina Butovskaya, Felipe~Nalon Castro, Hakan Cetinkaya,
  Diana Cunha, Daniel David, Oana~A. David, Fahd~A. Dileym, Alejandra del
  Carmen Domínguez~Espinosa, Silvia Donato, Daria Dronova, Seda Dural, Jitka
  Fialová, Maryanne Fisher, Evrim Gulbetekin, Aslıhan~Hamamcıoğlu Akkaya,
  Ivana Hromatko, Raffaella Iafrate, Mariana Iesyp, Bawo James, Jelena
  Jaranovic, Feng Jiang, Charles~Obadiah Kimamo, Grete Kjelvik, Fırat Koç,
  Amos Laar, Fívia de Araújo~Lopes, Guillermo Macbeth, Nicole~M. Marcano,
  Rocio Martinez, Norbert Mesko, Natalya Molodovskaya, Khadijeh Moradi,
  Zahrasadat Motahari, Alexandra Mühlhauser, Jean~Carlos Natividade, Joseph
  Ntayi, Elisabeth Oberzaucher, Oluyinka Ojedokun, Mohd Sofian~Bin Omar-Fauzee,
  Ike~E. Onyishi, Anna Paluszak, Alda Portugal, Eugenia Razumiejczyk, Anu
  Realo, Ana~Paula Relvas, Maria Rivas, Muhammad Rizwan, Svjetlana
  Salkičević, Ivan Sarmány-Schuller, Susanne Schmehl, Oksana Senyk,
  Charlotte Sinding, Eftychia Stamkou, Stanislava Stoyanova, Denisa Šukolová,
  Nina Sutresna, Meri Tadinac, Andero Teras, Edna Lúcia~Tinoco Ponciano, Ritu
  Tripathi, Nachiketa Tripathi, Mamta Tripathi, Olja Uhryn, Maria~Emília
  Yamamoto, Gyesook Yoo, and Jr. John D.~Pierce.
\newblock Preferred interpersonal distances: A global comparison.
\newblock {\em Journal of Cross-Cultural Psychology}, 48(4):577--592, 2017.

\bibitem{tancik2020fourfeat}
Matthew Tancik, Pratul~P. Srinivasan, Ben Mildenhall, Sara Fridovich-Keil,
  Nithin Raghavan, Utkarsh Singhal, Ravi Ramamoorthi, Jonathan~T. Barron, and
  Ren Ng.
\newblock Fourier features let networks learn high frequency functions in low
  dimensional domains.
\newblock {\em NeurIPS}, 2020.

\bibitem{tewari2020state}
Ayush Tewari, Ohad Fried, Justus Thies, Vincent Sitzmann, Stephen Lombardi,
  Kalyan Sunkavalli, Ricardo Martin-Brualla, Tomas Simon, Jason Saragih,
  Matthias Nießner, Rohit Pandey, Sean Fanello, Gordon Wetzstein, Jun-Yan Zhu,
  Christian Theobalt, Maneesh Agrawala, Eli Shechtman, Dan~B Goldman, and
  Michael Zollhöfer.
\newblock State of the art on neural rendering, 2020.

\bibitem{tewari17MoFA}
Ayush Tewari, Michael Zoll{\"o}fer, Hyeongwoo Kim, Pablo Garrido, Florian
  Bernard, Patrick Perez, and Theobalt Christian.
\newblock {MoFA: Model-based Deep Convolutional Face Autoencoder for
  Unsupervised Monocular Reconstruction}.
\newblock 2017.

\bibitem{Justus19}
Justus Thies, Michael Zollh\"{o}fer, and Matthias Nie\ss{}ner.
\newblock Deferred neural rendering: Image synthesis using neural textures.
\newblock {\em ACM Trans. Graph.}, 38(4), 2019.

\bibitem{Tran_2018_CVPR}
Luan Tran and Xiaoming Liu.
\newblock Nonlinear 3d face morphable model.
\newblock June 2018.

\bibitem{Shih-En19}
Shih-En Wei, Jason Saragih, Tomas Simon, Adam~W. Harley, Stephen Lombardi,
  Michal Perdoch, Alexander Hypes, Dawei Wang, Hernan Badino, and Yaser Sheikh.
\newblock Vr facial animation via multiview image translation.
\newblock {\em TOG}, 38(4), 2019.

\bibitem{Lior20}
Lior Yariv, Yoni Kasten, Dror Moran, Meirav Galun, Matan Atzmon, Ronen Basri,
  and Yaron Lipman.
\newblock Multiview neural surface reconstruction by disentangling geometry and
  appearance.
\newblock 2020.

\bibitem{Zhou_2019_CVPR}
Yuxiang Zhou, Jiankang Deng, Irene Kotsia, and Stefanos Zafeiriou.
\newblock Dense 3d face decoding over 2500fps: Joint texture \& shape
  convolutional mesh decoders.
\newblock June 2019.

\end{thebibliography}
}
\end{document}